\documentclass[11pt]{article}
\usepackage{times}
\usepackage{url}
\usepackage{latexsym}
\usepackage{graphics} 
\usepackage{multirow}
\usepackage[pdftex]{hyperref}
\usepackage[colorlinks]{}
\usepackage{color}
\usepackage{booktabs}
\usepackage{colortbl}
\usepackage{eacl2017}

\newcommand\tabhead[1]{\small\textbf{#1}}

\eaclfinalcopy 

\title{Argument Strength is in the Eye of the Beholder: \\ Audience Effects in Persuasion}

\author{Stephanie Lukin, Pranav Anand, Marilyn Walker and Steve Whittaker \\
  Computer Science, Linguistics and Psychology Depts. \\
  University of California, Santa Cruz \\
  Santa Cruz, Ca. 95064 \\
  {\tt slukin,panand,mawalker,swhittak@ucsc.edu}}

\date{}

\begin{document}
\maketitle
\begin{abstract}
Americans spend about a third
of their time online, with many participating in online conversations
on social and political issues. We hypothesize that social media
arguments on such issues may be more engaging and persuasive than traditional media summaries, 
and  that particular types of people may be more
or less convinced by particular styles of argument, e.g.  emotional
arguments may resonate with some personalities while factual arguments
resonate with others.  We report a set of experiments testing at large
scale how audience variables interact with argument style to affect
the persuasiveness of an argument, an under-researched topic within
natural language processing. We show that belief change is affected
by personality factors, with conscientious, open and agreeable people
being more convinced by emotional arguments.
\end{abstract}

\section{Introduction}
\label{intro-sec}

Americans spend a third of their online time on social media, with
many participating in online conversations about  education,
public policy, or other social and political issues.  Our hypothesis
is that online dialogs have important properties that  may make them a
useful resource for educating the public about such
issues.  For example, user-generated content might be more engaging and
persuasive than traditional media, due to the prevalence of emotional
language, social affiliation, conversational argument structure and
audience involvement. Moreover, particular types of people may be more
or less convinced by particular styles of argument, e.g.  emotional
arguments may resonate with some personalities while factual arguments
resonate with others.

\begin{table}[!thb]
\begin{center}
\begin{small}
\begin{tabular}{@{}p{0.5cm}|p{2.5in}@{}}\toprule
\multicolumn{2}{c}{\bf \cellcolor[gray]{0.9} {\bf Factual: Death Penalty}} \\ \midrule
{\bf Q1}: & I'm sure there have been more repeat murderers than innocent people put to death. As far as the cost goes, is that really an issue? Execution Room = \$10,000. Stainless Steel Table = \$2,000. Leather Straps = \$200. Lethal Injection Chemicals = \$5,000. Knowing this person will never possibly be able to kill again = PRICELESS \\
{\bf R1}:& Actually the room, straps, and table are all multi-use. And the drugs only cost Texas \$86.08 per execution as of 2002. \\ \toprule
\multicolumn{2}{c}{\bf \cellcolor[gray]{0.9} {\bf Emotional: Death Penalty}} \\ \midrule  
{\bf Q2}: & You mean, the perpetrator is convicted and the defender acquitted? Yes, that's the rule and not the exception. Notice here how no-one ended up dead, or even particularly seriously injured. Additionally the circumstances described are incredibly rare, that's why it makes the news.  \\
{\bf R2}: & The defender shouldn't even have been brought to trial in the first place. That doesn't make it any better. Somebody breaks into your home and threatens your family with rape and murder, they deserve serious injury at the very least. 
\\ \bottomrule
\end{tabular}
\end{small}
\end{center}
\vspace{-0.08in}
\caption{\label{sample-quote-response} Factual vs. Emotional dialog exchanges  
{\small \tt 4forums.com}. {\bf Q} = Quote, {\bf R} = Response.}
\vspace{-.1in}
\end{table}

For example, contrast the two informal dialogic exchanges about
the death penalty in Table~\ref{sample-quote-response} with the
traditional media professional summary in Table~\ref{death-procon}.  We might
expect the argument in 
Table~\ref{death-procon} to be more convincing, because it is carefully
written to be balanced and exhaustive \cite{ReedRowe04}. On the other hand, it 
seems possible that people find dialogic arguments
such as those in Table~\ref{sample-quote-response} more engaging and
learn more from them.  And indeed, about 90\% of the people in online
forums are so-called lurkers
\cite{Whittaker96,NonneckePreece00,Preeceetal04}, and do not post,
suggesting that they are in fact {\bf reading} opinionated dialogs
such as those in Table~\ref{sample-quote-response} for interest or
entertainment.

\begin{table}[th!]
\begin{small}
\begin{tabular}{@{}p{0.65cm}|p{6.35cm}@{}}\toprule
\multicolumn{2}{c}{\bf \cellcolor[gray]{0.9} Curated Summary: Death Penalty }   \\ \midrule  
{\bf PRO:} & Proponents of the death penalty say it is an important tool for preserving law and order, deters crime, and costs less than life imprisonment. They argue that retribution or "an eye for an eye" honors the victim, helps console grieving families, and ensures that the perpetrators of heinous crimes never have an opportunity to cause future tragedy.
 \\ \midrule
{\bf CON:} & Opponents of capital punishment say it has no deterrent effect on crime, wrongly gives governments the power to take human life, and perpetuates social injustices by disproportionately targeting people of color (racist) and people who cannot afford good attorneys (classist). They say lifetime jail sentences are a more severe and less expensive punishment than death. \\ \bottomrule
\end{tabular}
\end{small}
\vspace{-0.1in}
\caption{\label{death-procon} Traditional balanced summary of the death penalty
issue from {\small \tt ProCon.org}.}
\end{table}

Research in
social psychology identifies three factors that affect argument
persuasiveness \cite{PettyCacioppo86,PettyCacioppo88}.
\begin{itemize}
\item the {\sc argument} itself
\vspace{-.1in}
\item the {\sc audience}
\vspace{-.1in}
\item the {\sc source} of the argument
\end{itemize}

The {\sc argument} includes the content and its presentation,
e.g. whether it is a monolog or a dialog, or whether it is factual or
emotional as illustrated in Table~\ref{sample-quote-response} and
Table~\ref{death-procon}.  The {\sc audience} factor models people's
prior beliefs and social affiliations as well as innate individual
differences that affect their susceptibility to particular arguments
or types of arguments \cite{Anderson71,Davies98,Devineetal00,Pettyetal81}.
Behavioral economics research shows that the cognitive style of the
audience interacts with the argument's emotional appeal: emphasizing
personal losses is more persuasive for neurotics, whereas gains are
effective for extraverts \cite{Carveretal00,Mannetal04}.  The {\sc
  source} is the speaker, whose influence may depend
on factors such as attractiveness, expertise, trustworthiness or group
identification or homophily
\cite{EaglyChaiken75,Kelman61,bender2011annotating,LuchokMcCroskey78,Ludford04,mcpherson2001birds}.

We present experiments evaluating how properties of social media
arguments interact with audience factors to affect belief change. We
compare the effects of two aspects of the {\sc argument}: whether it
is monologic or dialogic, and whether it is factual or emotional. We
also examine how these factors interact with properties of the {\sc
  audience}. We profile audience {\bf prior beliefs} to test if more
neutral people are swayed by different types of arguments than people
with entrenched beliefs. We also profile the audience for Big Five
personality traits to see whether {\bf different personality types}
are more open to different types of arguments, e.g., we hypothesize
that people who are highly agreeable ({\sc A}) might be more affected by the
combative style of emotional arguments. We provide a new corpus for
the research community of audience personality profiles, arguments,
and belief change measurements.\footnote{{\tt 
    nlds.soe.ucsc.edu/persuasion\_persona}}

Audience factors have been explored in
social psychological work on persuasion, but
have been neglected in computational work, which has largely
drawn from sentiment, rhetorical, or argument structure models
\cite{habernal2016argument,ConradWiebe12,BoltuzicSnajder14,ChoiCardie08}. We
demonstrate that, indeed, undecided people respond differently to
arguments than entrenched people, and that the responses of undecided people
correlate with personality. We show that this holds across an array of
different arguments. Our research questions
are:
\begin{itemize}
\item Can we mine social
media to find arguments that change people's beliefs?
\vspace{-.1in}
\item Do different
argument types have different effects on belief change?
\vspace{-.1in}
\item Do
personality and prior beliefs affect belief change?
\vspace{-.1in}
\item Are
different personality types differently affected by factual
vs. emotional arguments?  
\end{itemize}

Our results show a small but highly reliable
effect that short arguments derived from online dialogs do lead people
to change their minds about topics such as abortion, gun control, gay
marriage, evolution, the death penalty and climate change. As
expected, opinion change is greater for people who are initially more
neutral about a topic, than those who are entrenched.  However
personality variables also have a clear effect on opinion change:
neutral, balanced arguments are more successful with all personality
types, but conscientious people are more convinced by dialogic
emotional arguments, and agreeable people are more persuaded by
dialogic factual arguments. We describe how we use plan these findings
to select and repurpose social media arguments to adapt them to people's
individual differences and thus maximize their educational impact.

\section{Related Work}
\label{relwork-sec}

Previous work on belief change has primarily focused on single,
experimentally crafted, persuasive messages, rather than exploring
whether user-generated dialogic arguments can be
repurposed to persuade.  Recently however several papers have begun to
investigate two challenges in argument mining: (1) understanding the
structure of an argument and extracting argument components
\cite{lippi2015context,nguyen2015extracting,stab2014annotating,lippi2015context,BiranRambow11};
  and (2) understanding what predicts the persuasiveness of
  web-sourced argumentative content
  \cite{habernal2016argument,ostendorf2016learning,wachsmuth2016using,habernalmakes16,Tanetal16}.

Tan et al.~\shortcite{Tanetal16} study belief change in the Reddit {\tt \small
  /r/ChangeMyView} subreddit (CMV), in which an original poster (OP)
challenges others to change his/her opinion.  They build logistic
regression models to predict argument success, identifying two
conversational dynamic factors: a) early potential persuaders are more
successful and b) after 4 exchanges, the chance of persuasion drops
virtually to zero. Linguistic factors of persuasive posts include: a)
dissimilar content words to the OP, b) similar stop words, c) being
lengthy (in words, sentences, and paragraphs), d) italics and
bullets. 
Finally, susceptibility to persuasion is correlated with
singular vs. plural first person pronouns, which the authors relate to
the personality trait of Openness to Experience. The CMV reddit offers
a unique window into how persuasion of self-declared open-minded
people occurs online. However, while Tan et al. find potential proxies
for personality traits, they cannot examine traits directly because
they do not have personality profiles as we do here. They also 
do not examine the effect of argument style as we do.

Recent work \cite{habernal2016argument,habernalmakes16} also examines
what makes an informal social media argument convincing. They have
created a new dataset of pairs of arguments annotated for which
argument is more convincing, along with the reasons given by
annotators for its convincingness.  They test several models for
predicting convincingness comparing an SVM with engineered linguistic
features to a BLSTM, with both models performing similarly. In
contrast to our experiments, they do not explore factors of the
audience or explicitly vary the style of the argument.


Previous work also tests the hypothesis that dialogic exchanges 
might be more engaging, in the context of expository or car sales dialog
\cite{Andreetal00,Lee10,Craigetal06,StoyanchevPiwek10}.
Work comparing monologic vs. dialogic modes of providing information
suggest that dialogs: (1) are more memorable and engaging, (2)
stimulate the audience to formulate their own questions, and (3) allow
audiences to be more successful at following communication
\cite{Leeetal98,Foxtree99,SuzukiYamada04,Driscolletal03,FoxtreeMayer08,Foxtree99,LiuFoxtree11}.

Other work \cite{Vydiswaranetal12} explores how user-interface factors
(e.g., number and order of argument presentation, whether and how
arguments are rated) affect how readers process arguments.
Several factors increased the
number of passages read, including explicitly presenting contrasting
viewpoints simultaneously. 
This exercise caused people
with strong beliefs (about the healthiness of milk) to moderate
their views after 20-30 minutes of concentrated study. 
We do not concentrate on interface factors, 
instead exploring how persuasiveness relates to audience
factors and argumentative style. 
Also  our experiments are run online
with hundreds of users, rather than as a controlled
study in the lab.


\section{Experimental Method}

Our experimental method consists of the following steps:

\begin{itemize}
\item Select user-generated dialogs with persuasive argument features from an online corpus of socio-political debates, exploring the role of {\bf affect} (Sec.~\ref{corpus-iac}).
\vspace{-0.08in}
\item Profile subjects for {\bf personality traits} and  {\bf prior beliefs} about socio-political issues (Sec.~\ref{corpus-pers}).
\vspace{-0.08in}
\item Expose subjects to user-generated,  factual vs. emotional
dialogic exchanges and compare the effects on belief change to
balanced, curated arguments (Sec.~\ref{corpus-belief}).
\vspace{-0.08in}
\item Conduct experiments to predict the degree of belief 
change as a function of prior belief, personality and type of argument.
\end{itemize}

The participants were pre-qualified using a reading comprehension task
that checked their responses against a gold standard to ensure that
they read the arguments carefully. Because we make many comparisons,
and our experiments are conducted at large scale, all of our results
incorporate Bonferroni corrections.

\subsection{Dialog Selection: Identifying Socio-Emotional Arguments}
\label{corpus-iac}

Our work requires a new experimental corpus that is sensitive to
readers' prior beliefs and personalities. We utilize online dialogs
from {\tt 4forums.com} downloaded from The
Internet Argument Corpus (IAC) \cite{Walkeretal12c}. The IAC contains
quote/response pairs of targeted arguments between two
people (Table~\ref{sample-quote-response}) on topics such as: death
penalty, gay marriage, climate change, abortion, evolution and gun
control.  Each argument is annotated to distinguish arguments making
strong appeals to emotional factors versus straightforwardly factual
arguments.

\begin{table}[t!hb]
\begin{center}
\begin{small}
\begin{tabular}{@{}p{0.5cm}|p{2.5in}@{}}\toprule
\multicolumn{2}{c}{\bf \cellcolor[gray]{0.9} {\bf Factual: Abortion}} \\ \midrule
{\bf Q3}: & Not only that, to suggest that untold numbers of women would seek illegal abortions is a question-begging claim that has no grounding in history, logic, or reason. It is an unfounded, unproven claim. It is a betrayal to sound judgment to make decisions based upon unfounded predictions into the future.  \\
{\bf R3}: & But it is based on history. There is plenty of history showing that women had illegal abortions. \\ \midrule
\multicolumn{2}{c}{\bf \cellcolor[gray]{0.9} {\bf Factual: Climate Change}} \\ \midrule

{\bf Q4}: & This is where the looney left gets lost. Their mantra is atmospheric CO2 levels are escalating and this is unquestionably causing earth's temperature rise. But ask yourself -- if global temperatures are experiencing the biggest sustained drop in decades, while CO2 levels continue to rise -- how can it be true? \\
{\bf R4}: & Because internal variability from the likes of ENSO, which can cause short term swings of a full degree C, easily swamp the smaller increase we'd expect from CO2 forcing. Easy. \\ \midrule
\multicolumn{2}{c}{\bf \cellcolor[gray]{0.9} {\bf Emotional: Abortion}} \\ \midrule  
{\bf Q5}: & Undesired first pregnancy is an acute problem for many girls who choose to go under the surgical knife, even though that often ends up with infertility, broken life etc. Dry fasting is an alternative to first pregnancy abortion. If applied, up to 2-3 months old embryo gets dissolved after 15-16 days of the fast. Plus, there is no 'christian' sin.  \\
{\bf R5}: & No Christian sin??? Other then the intent to kill and then doing so:p \\ \midrule
\multicolumn{2}{c}{\bf \cellcolor[gray]{0.9} {\bf Emotional: Gay Marriage}} \\ \midrule  
{\bf Q6}: &  Did anyone else expect anything less? These evil fundie christianists can have affairs, 2, 3, 4, or even 5 marriages yet gay people are a threat to marriage by wanting to get married. \\
{\bf R6}: & You hear that cry...allowing gays to marry will cause the downfall of civilization...but you never hear 'how' or 'why'? More Chicken Little \#\#\#\#. \\
\bottomrule
\end{tabular}
\end{small}
\end{center}
\vspace{-0.08in}
\caption{\label{more-fact-feeling} Factual vs. Emotional dialog exchanges.
{\bf Q} = Quote, {\bf R} = Response.}
\vspace{-.1in}
\end{table}

We selected a subset of extreme exemplars of factual ({\sc fact})
versus emotional ({\sc emot}) arguments, defined as Q/R pairs reliably
annotated to be at the extreme ends of the fact/emotion scale, i.e.
responses with an average $\ge 4$ annotation were considered factual,
and those whose annotation averaged $\le -4$ were considered emotional
on a scale of -5 to 5.  Table~\ref{sample-quote-response} illustrates
both factual (R1) and emotional (R2) arguments, with additional
examples for other topics in Table~\ref{more-fact-feeling}.

In the IAC, 95\% of the Q-R pairs are disagreements, the 
{\sc fact} and {\sc emot} datasets were selected to contain a similar
proportion. There was no correlation between agreement/disagreement
and emotionality (r $=0.07$, ns).

\subsection{Personality} 
\label{corpus-pers}

Personality is usually measured with a standardized survey that
calculates a scalar value for the five {\sc ocean} traits: opennness
to experience {\sc O}, conscientiousness {\sc C}, extraversion {\sc
  E}, agreeableness {\sc A}, and neuroticism {\sc N}
\cite{Goldberg90,Norman63} We first conducted an experiment to profile
the Big Five personality traits of 637 Turkers using the Ten Item
Personality Inventory (TIPI) \cite{Goslingetal03}.  The TIPI
instrument defines each person on a scale from 1 to 7 with 0.5
precision.  In order to guarantee reliablity of our results, we then
verified that our pre-qualified Turkers are representative of the
population as a whole, by comparing the means and standard deviations
of our sample of 637 Turkers with the national standards given in
Gosling et. al \shortcite{Goslingetal03}. Table~\ref{tab:distr} shows that our survey
means and standard deviations are very close to the national norms,
suggesting our sample is representative of the public in general, and hence
can be used to validate whether  social media arguments could fruitfully be
be used to educate the public.

\begin{table}[h!]
\begin{small}
\centering
\begin{tabular}{| c*{5}{|c} |}
\hline
&\tabhead{E} & \tabhead{A} & \tabhead{C} & \tabhead{N} & \tabhead{O} \\
\hline
Our survey&4.30&5.19&5.50&4.82&5.53 \\ 
Norms&4.44&5.23&5.4&4.83&5.38 \\
Our survey $\sigma$ &1.45&1.11&1.32&1.42&1.07 \\
Norms $\sigma$ &1.44&1.24&1.24&1.41&1.14 \\
\hline
\end{tabular}
\caption{TIPI $\sigma$ and mean from our personality survey compared to the normal distribution}
\label{tab:distr}
\end{small}
\vspace{-.2in}
\end{table}

\subsection{Prior Beliefs and Belief Change} 
\label{corpus-belief}

Previous research suggests that people who are entrenched about an
issue are unlikely to change their mind
\cite{Anderson71,Davies98,Devineetal00}, so we wanted to establish the
baseline beliefs of our pre-qualified Turkers before they had been
exposed to any arguments about a topic. We therefore collected each
Turker's initial stance on a topic, by asking them to answer a simple
{\bf stance question} with no context, for example: {\it Should the
  death penalty be allowed?}. Likert responses were recorded on a -5
to 5 slider scale with 0.01 degrees of precision, with labels on the
slider of ``Yes", ``No", or ``Neutral".

\begin{table}[th!]
\begin{small}
\begin{tabular}{@{}p{0.65cm}|p{6.35cm}@{}}\toprule
\multicolumn{2}{c}{\bf \cellcolor[gray]{0.9} Curated Summary: Abortion }   \\ \midrule  
{\bf PRO}: &  Proponents, identifying themselves as pro-choice, contend that abortion is a right that should not be limited by governmental or religious authority, and which outweighs any right claimed for an embryo or fetus. They argue that pregnant women will resort to unsafe illegal abortions if there is no legal option.  \\ \midrule
{\bf CON}:  & Opponents, identifying themselves as pro-life, assert that personhood begins at conception, and therefore abortion is the immoral killing of an innocent human being. They say abortion inflicts suffering on the unborn child, and that it is unfair to allow abortion when couples who cannot biologically conceive are waiting to adopt. \\ \bottomrule
\end{tabular}
\end{small}
\vspace{-0.1in}
\caption{\label{abortion-procon} Traditional balanced summary of ``Should abortion be legal?''  from {\small \tt ProCon.org}.}
\end{table}

Our goal is to compare the belief change that results from
social-media dialogs with the belief change from
professionally-curated monologs. We selected the balanced, monologic,
argument summaries from the website {\tt ProCon.org} (in
Table~\ref{death-procon} with an additional example in
Table~\ref{abortion-procon}). The arguments from {\tt ProCon.org} are
very high quality, and produced by domain experts.

After probing initial beliefs, we presented participants with one of
the three different argument types to test their affect on belief
change: a Curated Monolog ({\sc mono}) (Table~\ref{death-procon}), an
emotional argument ({\sc emot}) (R2 in
Table~\ref{sample-quote-response}), or a factual argument ({\sc fact})
(R1 in Table~\ref{sample-quote-response}).  After each person read one
of these three types of arguments, we retested their reactions to the
original stance question, while viewing the argument.  Responses were
again recorded on a -5 to 5 slider scale with 0.01 degrees of
precision, with labels on the slider of "Yes", "No", or "Neutral".  We
computed belief change by measuring differences in stance before and
after reading each argument.  We created 20 HITs on Mechanical Turk
for this task, with 5 items per hit.

\section{Experimental Corpus Results}

\subsection{Entrenchment and Belief Change}
Our first question is  whether our method changed participant's
beliefs.  Table ~\ref{tab:beliefchange_means_stddev} shows belief
change as a function of argument type: monologs ({\sc mono}), factual ({\sc fact})
and emotional ({\sc emot}). Belief change occurred for all argument types:
and the change was statistically significant as measured by paired t-tests
($t_{(5184)}$ = 38.31, p \textless 0.0001). 
This
confirms our hypothesis that social media can be mined for persuasive
materials. In addition, all three types of arguments independently led
to significant changes in belief.\footnote{(For {\sc mono}, $t_{(3184)}$
= 32.65, p \textless 0.0001, for {\sc fact}, $t_{(1019)}$ =
14.81, p \textless 0.0001, For {\sc emot}, $t_{(979)}$ =
14.35, p \textless 0.0001).}

\begin{table}
\begin{small}
\centering
\begin{tabular}{|c*{2}{|c}|c|}
\hline
& \tabhead{N} & \tabhead{Mean change} & \tabhead{$\sigma$ change} \\
\hline
{\sc mono} entrenched & 1826 & 0.50 & 1.09 \\
{\sc mono} neutral & 1359 & 0.62 & 0.71 \\
{\sc fact} entrenched & 258 & 0.27 & 0.79 \\
{\sc fact} neutral & 202 & 0.39 & 0.55 \\
{\sc emot} entrenched & 213 & 0.35 & 0.87 \\
{\sc emot} neutral & 187 & 0.37 & 0.54 \\
\hline \hline
ALL entrenched & 2951 & 0.43 & 1.00 \\
ALL neutral & 2234 & 0.51 & 0.65 \\ 
\hline
\end{tabular}
\caption{Means and $\sigma$ for belief change for neutral and entrenched participants presented with {\sc mono}, {\sc fact}, or {\sc emot} argument types. Neutrals show more belief change, and all argument types significantly affect beliefs}
\label{tab:beliefchange_means_stddev}
\end{small}
\vspace{-.2in}
\end{table}

One of the strongest theoretical predictions 
is that people with entrenched beliefs about an issue are less
likely to change their mind when provided new information about
that issue.  Table~\ref{tab:beliefchange_means_stddev}
shows the relationship between initial beliefs and extent of belief
change. We defined people as having more entrenched initial beliefs 
if their response to the initial stance question was within 0.5 points of 
the two ends of the scale, i.e. (1.0-1.5) or (4.5-5.0), indicating an extreme initial view.  

We tested whether people who were more entrenched initially showed
less change than those who were initially more neutral.  We conducted
a 2 Initial Belief (Entrenched/Neutral) X 3 Argument Type ({\sc
  mono}/{\sc emot}/{\sc fact}) {\sc anova}, with Belief Change as the
dependent variable, and Initial Belief and Argument Type as between
subjects factors. Again, as expected, initially Entrenched people
showed less change (M = 0.43) than those who began with
Neutral views (M = 0.51), {\sc anova} ($F_{(1,5179)}$=5.97, p =
0.015).

\subsection{Argument Type and Belief Change} 

We wanted to test whether the engaging, socially interesting, dialogic materials of {\sc emot} and {\sc fact} might promote more belief
change than balanced curated monologic summaries. We tested the differences between
argument types, finding a main effect for argument type
($F_{(2,5179)}$=31.59, p \textless 0.0001), with Tukey post-hoc tests
showing {\sc mono} led to more belief change than both {\sc emot} and {\sc fact}
(both p \textless 0.0001), but no differences between {\sc emot} and {\sc fact}
overall across all subjects (See
Table~\ref{tab:beliefchange_means_stddev}). Finally there was no
interaction between Initial Belief and Argument Type
($F_{(2,5179)}$=1.25, p \textgreater 0.05): so although neutrals show
more belief change overall, this susceptibility does not vary by argument type. 

\section{Predicting Belief Change}

Our results so far show that our arguments changed people's beliefs as
a function of their prior beliefs and argument type. However we aim to
automatically {\bf predict} belief change, and 
hypothesize that knowing a person's {\bf personality} in combination
with their {\bf prior beliefs} will allow us to select 
social-media arguments that are more persuasive
{\bf for a particular individual}.

Thus, we vary whether providing a learner with features about a
person's personality improves performance for predicting belief
change, when compared with providing information about degree of
entrenchment alone.  We use different representations for personality
and prior beliefs as features,  the raw score from the Likert slider for
belief change and the TIPI score, as well as normalizations of the raw
scores according to the distributions per topic, and finally
categorical binning of the transformed scores.

\vspace{-0.05in}
\subsection{Feature Development and Selection}
\vspace{-0.05in}
\label{sub:features}

New features were created by computing the z-transformation score from
the raw prior beliefs and personality traits scores.  Applying
Equation~\ref{ztrans} to the raw data creates a normal distribution
where the new mean is 0 and the standard deviation is 1.
For prior beliefs, $x_i$ is an individual prior belief for a particular topic,
$\bar{x_i}$ is the mean, and $\sigma_i$ is the standard deviation for
the particular topic.

\begin{equation}
\frac{x_i - \bar{x_i}}{\sigma_{x_i}}
\label{ztrans}
\end{equation}

Categorical bins are derived from the transformed scores to describe
the direction of the belief change by comparing prior and final
recorded beliefs.  The belief change is positive or negative depending
upon where the Turkers rate themselves on the belief scale, moving
more towards one side (1) or the other (5). Next, to control for
variance, we apply a z-transformation on change scores to create a
normal distribution. We classify the resulting distribution into three
bins: Low, Medium, and High. The interpretation of what stance the Low
and High bins represent is strictly topic dependent.  The Medium bin
consists of z-transformation values between -1 and 1. These are the
people whose belief change is less than one standard deviation from
the transformed mean. The Low bin contains z-transform scores of less
than -1 and translates to belief changes of a large magnitude (more
than a standard deviation from the mean) in a negative direction,
where again, the meaning of ``negative'' is dependent upon how the
question was framed. The High bin contains z-transformation scores of
greater than 1 and translates to belief changes of a large magnitude
in a positive direction.  For example, the stance question {\it
  Should the death penalty be allowed?'} has ``no'' at the -5 end and
``yes'' at the +5 end of the likert scale. A Low bin is indicative as
moving in the direction of the ``no'' stance and High towards the
``yes'' stance.

Bins were also derived for the personality
traits, e.g.  for Openness, the High bin indicates someone
who is very open, the Medium bin is average, and the Low bin is
not open at all.

Finally, a binary feature was created to represent how entrenched an
individual is in a particular topic. This feature is based on the raw
prior belief score and is True if the prior belief score is within 0.5
points of either end point on the stance scale. This feature is
different from the prior belief bins because this entrenchment feature
groups together people who are in the extremes on both sides of the
stance scale, while the prior belief bins distinguishes between the
two ends.

We created a development set using data from a prior Mechanical Turk
experiment which had 20 HITs, 5 questions per HIT, and 20 people who
completed each HIT. In the same manner as the {\sc fact} and {\sc
  emot} HITs, these Turkers (whose personality was already
profiled) were asked about their prior beliefs about a topic, then
presented with a factual or emotional argument. But in this case they
were asked to rate the strength of the argument rather than to report
their belief about the topic.  We then identified the combination of
features that best predicted argument strength in this development
data, and then used this feature set for the belief change experiments
below. Turkers who participated in this initial study did not
participate in the belief change study and vice versa.

Results on the development set showed that the z-transformation scores
for prior belief and personality performed better than the raw scores
and bins. On the other hand, the belief change feature was most
effective when represented as a categorical variable via binning and
directionality. We also found that it is better to have {\bf both} the
z-transformed prior belief feature and the entrenchment feature. 
Thus our experiments below use these
feature representations.

We test on three different datasets: {\sc mono}, {\sc fact} and {\sc
  emot}, to elicit responses from reading the monologic summaries, and
the factual and emotional dialogic arguments.  The {\sc fact} and {\sc
  emot} datasets have specific information in terms of scalar values
about their degree of factuality or emotionality, on a scale of -5,+5
and a feature with this value is created for these datasets derived
from the crowdsourced Turker judgments about the degree of
Fact/Emotion in a Q/R pair, as described earlier. The monologic
summaries ({\sc mono}) are assumed to be neutral and are not assigned
a value for degree of factuality or emotionality.


\subsection{Belief Change Experimental Results}
\label{bel-change-exp}

\begin{table*}[th!]
\begin{small}
\begin{center}
\begin{tabular}{|c*{6}{|c} |}
\hline
\tabhead{row \#} & \tabhead{Dataset} & \tabhead{TIPI}  & \tabhead{Accuracy} & \tabhead{Precision} & \tabhead{Recall} & \tabhead{F1} \\
\hline \hline
1 & \bf Baseline &  & 33\% & & & \\
\hline \hline
2 & {\sc mono} & None 		&	57\%	&	0.50	&	0.58	&	0.51	\\
3 & &	{\bf Open}		&	58\%	&	0.52	&	0.59	&	\bf 0.52	\\
4 & &	Conscientious& 	58\% 	&	0.51	&	0.58	&	0.51	\\
5 & &	Extrovert			& 	58\% 	&	0.49	&	0.57	&	0.49	\\
6 & &	Agreeable		& 	58\% 	&	0.52	&	0.57	&	0.50	\\
7 & &	Neurotic			& 	57\% 	&	0.49	&	0.55	&	0.47	\\
8 & &	\bf All				& 	58\% 	&	0.52	&	0.58	&	\bf 0.52	\\
\hline \hline
9 & {\sc fact}	& None				&	49\% 	&	0.47	&	0.47	&	0.46	\\
10 & &	 Open		& 	46\% 	&	0.45	&	0.47	&	0.46	\\
11 & &	Conscientious		& 	48\% 	&	0.48	&	0.45	&	0.46	\\
12 & &	Extrovert			& 	48\% 	&	0.46	&	0.45	&	0.44	\\
13 & &	{\bf Agreeable}		& 	51\% 	&	0.52	&	0.49	&	\bf 0.49	\\
14 & &	Neurotic			& 	47\% 	&	0.45	&	0.44	&	0.43	\\
15 & &	{\bf All}				&     50\% 	&	0.49	&	0.50	&	\bf 0.49	\\
\hline \hline
16 & {\sc emot} &	None&	53\% 	&	0.42	&	0.52	&	0.44	\\
17 & &	{\bf Open}	&	56\% 	&	0.54	&	0.53	&	\bf 0.51	\\
18 & &	{\bf Conscientious}	&	53\% 	&	0.49	&	0.51	&	\bf 0.48	\\
19 & &	Extrovert	&	49\% 	&	0.43	&	0.47	&	0.44	\\
20 & &	{\bf Agreeable}	&	52\% 	&	0.48	&	0.50	&	\bf 0.48	\\
21 & &	Neurotic	&	53\% 	&	0.43	&	0.51	&	0.44	\\
22 & &	{\bf All}	&	56\% 	&	0.55	&	0.57	&	\bf 0.56	\\
\hline
\end{tabular}
\caption{Predicting Belief Change with Naive Bayes for Three Data
  Sets: Statistics are based on 10-fold cross validation. Row numbers 
provided to reference particular results in the text.}
\label{tab:belief_change}
\end{center}
\end{small}
\end{table*}

Our dataset consists of 5185 items, with  3185 responses to
the balanced {\sc mono} summaries, 1020 responses to {\sc fact}, and
980 responses to {\sc emot}.  We first applied 10-fold
cross-validation with Naive Bayes, Nearest Neighbor, AdaBoost, and JRIP,
from the Weka toolkit \cite{WittenFrank05}.  Overall, Naive Bayes had
the most consistent scores with our feature sets, thus we only report
Naive Bayes experimental results below.

Seven feature sets were created for each of the three \{{\sc mono}, {\sc
  fact}, {\sc emot}\} datasets.  {\it None} feature sets are 
the no-personality baseline within each dataset. The
baseline features contain {\bf no information} about the personality
of the unseen human subjects. We use the \{{\sc mono}, {\sc fact}, {\sc
  emot}\}+None feature sets for testing our hypothesis that
personality affects belief change, and {\bf our ability to predict
  belief change} using personality features.  {\it All} feature sets
have information about {\bf all} of the human subjects' personality
traits as 5 distinct features.  The remaining five {\it \{O,C,E,A,N\}}
feature sets examine the effect of providing information to the
learner about personality using only {\bf one personality trait at a
  time}, in order to determine if any personality trait is having
a larger impact for belief
change prediction.

Table~\ref{tab:belief_change} summarizes our key results, reporting
accuracy, precision, recall, and F1 for predicting belief change as a
discrete bin, Low, Medium, and High. We balanced each dataset to contain
the same number of instances in bins, thus the accuracy for
majority classification is 33\% (Row 1).

After running Naive Bayes over all feature sets in the three datasets,
we compared the experimental classifier performance of {\it All} and
{\it \{O,C,E,A,N\}} against the None baselines using a Bonferroni
corrected t-test for F1 measure. Using statistical {\sc anova} tests
that control for pre and post test sample variance, we found small but
highly reliable effects. We show all of our results, but focus our
discussion below on statistically significance differences in F1. We
boldface personality feature sets in Table~\ref{tab:belief_change}
that are statistically significant when comparing \{{\sc mono}, {\sc
  fact}, {\sc emot}\}+None with the other feature sets in the group.

The effect of argument alone (without personality information) can be
seen by the no-personality baseline for each argument type, where we
exclude personality information (\{{\sc mono}, {\sc fact}, {\sc
  emot}\}+None).  All these feature sets perform above the baseline of
33\% (Row 1). This supports the results of
our prior {\sc anova} testing over all subjects for belief change, and
shows that the argument itself partially predicts belief change.

However, more interestingly, Table~\ref{tab:belief_change} also shows
that providing the learner with information about personality
consistently improves the ability of the learner to predict belief
change. For all types of arguments, ie. the neutral, monologic
summaries and the factual and emotional dialogs, the feature sets
without any information about the personality traits of the unseen
human subjects perform significantly worse than the feature sets that
contain all five personality traits. {\sc mono}+None compared to {\sc
  mono}+All (rows 2 and 8 respectively) show a slight but significant
increase in F1 from 0.51 to 0.52 (using a paired t-test on 10 fold
cross validation scores, (p = .001)). Similarly, {\sc fact}+None
versus {\sc fact}+All (rows 9 and 15) shows a significantly greater
increase in F1: from 0.46 to 0.49 (p = .0002), as does {\sc
  emot}+None versus {\sc emot}+All (rows 16 and 22) with F1 increasing
from 0.44 to 0.56 (p = .00001). This confirms that the personality traits
improves a model's ability to predict belief change in unseen human
subjects.


Next we compared the effects of providing the learner with information
about each individual personality feature {\bf in isolation} by
comparing \{{\sc mono}, {\sc fact}, {\sc emot}\}+None with individual personality
factors. For {\sc mono}, we found that adding personality information about
Openness to Experience ({\sc mono}+O, row 3) improved F1 from 0.51 to 0.52 compared with a
no-personality baseline (p = .0006). 
This suggests that open people are more persuaded by 
balanced monologic arguments. 

A more interesting result is that Openness to Experience ({\sc
  emot}+O, row 17) was also important for Emotional arguments,
increasing F1 from 0.44 to 0.51 (p = .00001).  In contrast, Openness
had no effect for Factual arguments (Row 10) (p $>$ 0.05).  Models for
predicting belief change for Emotional arguments also benefit from
information about Conscientiousness and Agreeableness.
Row 17 ({\sc emot}+O), Row 18 ({\sc emot}+C) and Row 20
({\sc emot}+A) all show significant differences in F1, with {\sc
  emot}+O  better than {\sc emot}+None (p = .00001), {\sc
  emot}+C  better than {\sc emot}+None (p = .00001) and
{\sc emot}+A  better than {\sc emot}+None (p = .0001).

Information about Agreeableness  also improves
the quality of the belief change models for the factual dialogs
({\sc fact}+A, row 5) with an increase in F1 from 0.46 baseline to
0.49 (p = .004), suggesting that people who are more Agreeable are
more influenced by factual arguments.  This confirms one of our
initial hypotheses that Agreeable people would be more sensitive to
the fact/emotional dimension of arguments because of their desire to
either avoid conflict (highly Agreeable people) or to seek conflict
(Disagreeable people).

\section{Conclusions}
\label{conc-sec}

To the best of our knowledge we are the first to examine the interaction
of social media argument types with audience  factors. Our contributions are:
\begin{itemize}
\item A new corpus of personality information and belief change in
  socio-political arguments;
\vspace{-0.08in}
\item A new method for   identifying and deploying social media content 
to inform and   engage the public about important social and political
  topics;
\vspace{-0.08in}
\item Results showing at scale (hundreds of users) that 
we can mine  arguments from online discussions to change people's beliefs;
\vspace{-0.08in}
\item Results showing that different types of arguments have
  different effects: while balanced monologic summaries led to the greatest
  belief change, 
  socio-emotional online exchanges 
also caused changes in belief.
\end{itemize}

Although our short question/response pairs did not induce as much
belief change as the curated balanced monologs, we believe that these
are striking results given that the materials we extracted from online
discussions are not balanced or professionally produced, but instead
are simple fragments extracted from online discussions.

Further, confirming prior work on persuasion
\cite{EaglyChaiken75,Kelman61,Pettyetal81}, we found that these
effects depend on audience characteristics. As expected, belief
depended on the strength of prior beliefs so that initially neutral
people were more likely to be persuaded than entrenched individuals,
regardless of the type of argument. Again supporting our predictions,
argument effectiveness depended on personality type. People who are
Open to Experience were influenced by balanced and emotional
materials. In contrast, Agreeable people are most affected by factual
materials. Emotional arguments had very different effects from factual
and balanced monologs: Openness is important but so too are
Conscientiousness and Agreeableness. 

How can we explain this? People
who are more Open are typically receptive to new ideas.
But our results for emotional arguments also show that
Conscientious people change their views when presented with emotional
arguments, possibly because they are careful to process the arguments
however expressed. And Agreeable people may
also be motivated to change belief by emotional arguments because they
are less likely to be influenced by personal feelings. 

Our results have numerous implications that suggest further technical
experimentation. The fact that we can induce belief change by
extracting simple discussion fragments suggests that belief change can
be induced without the application of sophisticated text processing
tools. While our results for balanced monologs suggest that summaries
increase belief change, summary tools for such arguments are still
under development \cite{Misraetal15}.  However, perhaps 
high quality summaries may not be needed if compelling argument
fragments can be automatically extracted
\cite{Misraetal16,subba2007automatic,nguyen2015extracting,Swansonetal15}.

Our work also suggests the importance of personalization for
persuasion: with different personality types being open to different
styles of argument. Future work might be based on methods for
profiling participant personality from simple online behaviors
\cite{di2013detecting,liu2016buy,pan2014pplum,Ducheneautetal11}, or from user-generated
content such as first-person narratives or conversations \cite{MairesseWalker06a,MairesseWalker06b,rahimtoroghi2016learning,Rahimtoroghietal14}.  We could then select
personalized arguments to meet a participant's processing style.

While here we used crowdsourced judgments to select arguments of
particular types. Elsewhere, we present algorithms for
automatically identifying and bootstrapping arguments
with different properties. We have methods to extract 
arguments that represent different stances on an issue \cite{Misraetal16b,Anandetal11,Sridharetal15,Walkeretal12a,Walkeretal12b}, as well as argument
exchanges that are agreements vs. disagreements \cite{MisraWalker13}, factual
vs. emotional arguments \cite{Orabyetal15}, sarcastic and not-sarcastic arguments, and
nasty vs. nice arguments \cite{Orabyetal16,LukinWalker13,Justoetal14}. 

An open question is to whether these effects are long term. Our
approach limits us to examining belief change during a single session
for practical reasons; long-term cross-session comparisons lead to
significant participant retention issues. 

Our results also suggest new empirical and theoretical methods for
studying persuasion at scale. Only recently have studies of persuasion
moved beyond small scale lab studies involving simple single arguments
\cite{habernal2016argument,habernalmakes16,Tanetal16}.  Our research
also suggests new methods and tools for larger scale studies of
persuasion. While care must be taken in deploying these results,
studies of juries and other decision making bodies suggest that
exposure to a diversity of opinions and minority views are very
important to countering extremism and understanding the issues at
stake \cite{Devineetal00,Ludford04}. The ability to repurpose the huge
number of varied opinions available in social media sites for
educational purposes could provide a novel way to expose people to a
diversity of views.

\end{document}